# ANALYSIS OF THE SYNERGY BETWEEN MODULARITY AND AUTONOMY IN AN ARTIFICIAL INTELLIGENCE BASED FLEET COMPETITION

**Xingyu Li, PhD[1], Mainak Mitra, PhD[1], Bogdan I. Epureanu, PhD[1]**

[1] Department of Mechanical Engineering, University of Michigan, Ann Arbor, MI

## ABSTRACT

*A novel approach is provided for evaluating the benefits and burdens from vehicle modularity in fleets/units through the analysis of a game theoretical model of the competition between autonomous vehicle fleets in an attacker-defender game. We present an approach to obtain the heuristic operational strategies through fitting a decision tree on high-fidelity simulation results of an intelligent agent-based model. A multi-stage game theoretical model is also created for decision making considering military resources and impacts of past decisions. Nash equilibria of the operational strategy are revealed, and their characteristics are explored. The benefits of fleet modularity are also analyzed by comparing the results of the decision making process under diverse operational situations.*

## 1. INTRODUCTION

Military vehicles encounter diverse operational environments and use-case scenarios which demand flexibility and diversity in functional requirements of vehicles in a fleet. Some vehicles may also have specialized tactical functions and significant amounts of expendable resources. Changes in tactical needs and technological updates make the adaptation of such fleets and their immediate reuse after a mission difficult [1]. The US Army aims to keep such scenarios to a minimum, and to increase efficiencies in operations, by requiring that fleets of vehicles be re-utilizable across a variety of mission scenarios.

One option to achieve such flexibility is the introduction of vehicle modularity [2], wherein vehicles are built from swappable components known as modules. These modules include physical variants with varied functionalty that enable fast upgrades and efficient adaptation. Design of such modular vehicles has been explored extensively in the past, with various proposals addressing diverse military requirements, such as the armored vehicle family [3] and future combat systems [4]. Main advantages of modular vehicle fleets are their cost-effectiveness as well as their flexibility in operations achivable through plug-in/pull-out actions on the base and possibly on the battlefield [5]. The capability of performing these assembly, disassembly and reconfiguration actions (ADR) distinguishes modular fleets from conventional fleets and enable adjustments in configuration in reacting to demands in real-time.

While modularity offers additional flexibility in fleet operation, its advantages may be squandered by inadequate fleet operation strategies. The management of a modular fleet presents new challenges, especially considering the highly uncertain demands created by intelligent adversaries. In the past, operation management techniques have been applied to study the reduction of operational and supply cost of modular fleets by using an agent-based approach [6,7]. However, those studies made simplifying assumptions including deterministic values of demands and stationary adversarial behavior.

Relaxing these constraints, Li & Epureanu [8] proposed an attacker-defender game to simulate the competition between two adversarial and intelligent military forces. The benefits and burdens of modularity are revealed by simulating an attacker-defender game wherein the modular fleet operated against a conventional fleet. The modular and conventional fleets are randomly designated as attacker or defender. The objective of the defender is the satisfaction of the delivery of a convoy with consideration of possible attributes reduction due to vehicle damage. The goal of the attacker is to disrupt the defender from satisfying the demands. The modular fleet showed a better performance in the intelligent competition. With additional operational flexibility from ADR actions, the modular fleet exhibited a better adaptability and was less predictable because of its added operational flexibility. However, because of the complexity of the model, strategic interactions between decision makers were not obvious. Also, the equilibrium strategies and their evolution at different stages of the game was not identified.

In this paper, we address these issues through a data-driven appoach. The simulation data from the attacker-defender game is fit into two approximated models, namely a decision tree (DT) model and a game theoretical model, for gaining insights into the benefits and burdens of fleet modularity. DTs have been used in the past in diverse areas such as expert systems, signal classification and decision analysis. DTs have also been used in the management of military operation planning [9], software systems [10], and predition of adversarial actions [11]. The most important feature of a DT is the capacity to break the comlex decision-making process into a collection of simpler decisions to provide a interpretable solution. Through fitting of competition history between intelligent agents, a DT is used as a tool to analyze and reveal the popular heuristics of the intelligent entity.

Game-theory is an analysis technique for describing strategic interactions and their likely outcomes between multiple players. The games can be analyzed to find the equilibrium points and suggest the beneficial strategies. Game theoretical models have been widely used to simulate a large variety of military scenarios: information warfare [12], cyber attacks [13], submarine war [14], and sensor networks [15]. There are also several studies focused on attacker-defender games that consider resource-dependent strategies. For example, Powell [16] used a game-theoretical approach to find the defender's resource allocation strategy for protecting resources from being destroyed, which leads to the Bayesian-Nash equilibrim. Hausken and Zhuang [17] formulated an attacker-defender game where a defender decides to use the resources to protect themselves or attack the adversary for multiple time points. In this study, a game theoretical model is formualted to fit the simulation results from our previous model [8] and shows the impact of length of game in the decision making.

In section 2, we first brief our previous approach of formulating an attacker-defender game between two intelligent and adversarial forces. In section 3, the DT model is introduced for mining the heuristics of fleet operation. In section 4, a game theoretical model is built to find the equilibrium strategies in a multi-period game. In section 5, we draw conclusions and discuss a prospective future research direction.

## 2. Intelligent Agent-Based Model

Military demands are time-varying and highly uncertain because commanders react to adversarial actions. To capture these characteristics, an attacker-defender game is created between two hostile vehicle fleets assuming all vehicles are autonomous. The game is formulated as a transportation mission with uncertain assaults from the adversary. Demands are stochastically generated at battlefields specifying the requirements for the delivery, i.e., firepower, capacity. The fleet to satisfy the demands becomes the defender with the goal of delivering a convoy with sufficient attributes that satisfy the requirements. The other fleet becomes the attacker with the goal of disrupting the defender by an assault convoy. Vehicles in convoys are stochastically damaged during confrontations, which reduces the magnitude of the attributes of vehicles and convoys.

To increase the probability of winning a mission, the defender can infer the possible adversarial attack strategy to prepare sufficient vehicles in the convoy in case of possible damage during the confrontation. Denote the safety level for the defender/attacker as the ratio between actual attributes and demand requirements. 10 strategies are created for both the attacker and the defender, as shown in Table 1.

Concerning the efficiency and fidelity of the model, agents with different functionality are created to collaboratively make the operational and dispatch decisions. Following the previous work [8], three types of agents are defined as:

- **Inference agent**: to analyze adversarial future actions based on historical records

**Table 1:** Available strategies for the attacker and the defender and the corresponding ranges of safety levels

| Attack Strategy | 1 | 2 | 3 | 4 | 5 | 6 | 7 | 8 | 9 | 10 |
|---|---|---|---|---|---|---|---|---|---|---|
| Firepower | $[0, 0.5)$ | $[0.5, 1)$ | $[1, 1.5)$ | $[1.5, 2)$ | $[2, 2.5)$ | $[2.5, 3)$ | $[3, 3.5)$ | $[3.5, 4)$ | $[4, 4.5)$ | $[4.5, \infty)$ |
| Defense Strategy | 1 | 2 | 3 | 4 | 5 | 6 | 7 | 8 | 9 | 10 |
| Firepower | $[0, 1)$ | $[1, 1.5)$ | $[1, 1.5)$ | $[1, 1.5)$ | $[1.5, 2)$ | $[1.5, 2)$ | $[1.5, 2)$ | $[2, \infty)$ | $[2, \infty)$ | $[2, \infty)$ |
| Capacity | $[0, 1)$ | $[1, 1.5)$ | $[1.5, 2)$ | $[2, \infty)$ | $[1, 1.5)$ | $[1.5, 2)$ | $[2, \infty)$ | $[1, 1.5)$ | $[1.5, 2)$ | $[2, \infty)$ |

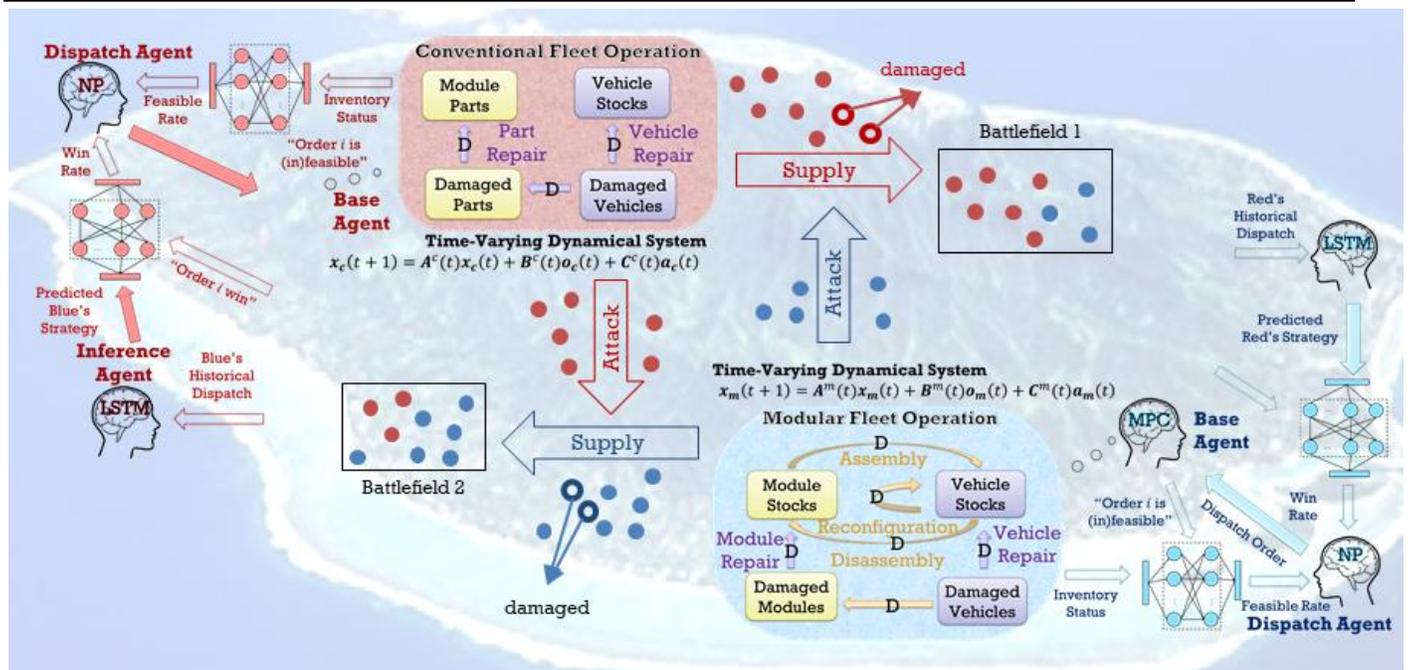

**Figure 1:** The attacker-defender game between the modular fleet and the conventional fleet

- **Dispatch agent**: to determine optimal dispatch orders based on feasibility and success rate
- **Base agent**: to schedule ADR actions and repair actions to fulfill dispatch orders

The attacker-defender game is used to simulate the competition between a conventional fleet and a modular fleet with no resupply. All damaged vehicles and components can only be reused after a long repair (recovery) time. The simulation is separated into two stages: (1) the stochastic stage, where each fleet randomly selects the dispatch strategy, and (2) the intelligent stage where each fleet makes decisions through and artificial intelligence analysis that uses models trained by data collected from the stochastic stage.

Downtime is considered in the attacker-defender game, which includes the time for ADR actions and vehicle recovery. In this study, the vehicle downtime is calculated by summing the processing time required for each action. For example, the time to reconfigure a vehicle from one type to another is the sum of time needed to disassemble the modules from old types and the time to assemble the new modules required by the new types. In this study, the time for module assembly is 1 hour, for module disassembly is 0.5 hours, and for module recovery is 10 hours. Previous studies have shown that different values of downtime change the modular fleet behaviors [7]. In this study, changes in the downtime will also impact the simulation outcomes of the attacker-defender game, which change the analysis results in this study.

Previous work has shown the benefits and burdens of modularity in terms of the win rate and the ability to avoid being inferred. In this study, we analyze those results in a new perspective to gain actionable insights from the data through two simplified models with high interpretability: (a) a DT for extracting heuristic rules from high-fidelity data, and (b) a game theoretical model for discovering game Nash equilibria.

## 3. Decision Tree (DT)

DTs are popular decision analysis tool which are used here to interpret the decision making process of each fleet. A DT is a flowchart like tree structure. Each node of the tree indicates a certain operational situation. The value represents the number of winning cases (1) and losing cases (0) in that situation. Each edge of the tree denotes an operational condition, e.g., attack strategy $\leq$ strategy 3. A node with two branches creates a single-stage classifier, which provides an intuitive comparison of the payoffs by making different decisions in the given situation.

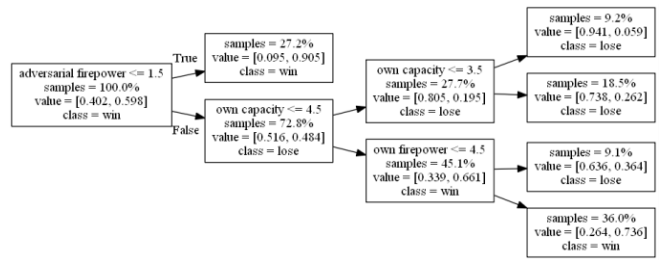

**Figure 2.1:** DT of the modular fleet engaged as the defender.

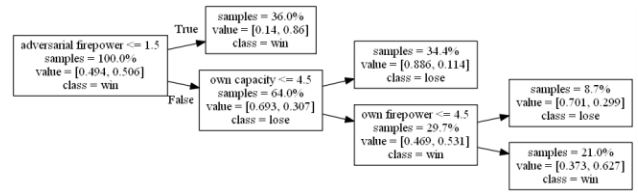

**Figure 2.2:** DT of the conv. fleet engaged as the defender.

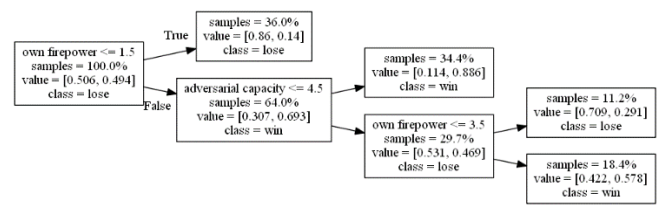

**Figure 2.3:** DT of the modular fleet engaged as the attacker.

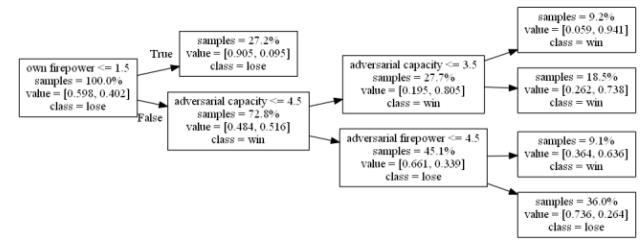

**Figure 2.4:** DT of the conv. fleet engaged as the attacker.

The DT is trained by partitioning the training set by the operational condition. The goal of each partition is to increase the purity of separated subsets. The partition ends once the threshold of purity is reached. The purity is measured by entropy, a value that describes the chaos of a set, expressed by

$$\text{entropy} = \sum_{i=0}^{i=1} -p_i \log_2 p_i.$$

where, $p_i$ is the probability of winning ($i = 1$) and losing ($i = 0$). The condition that minimizes the total entropy of the subset is selected for partition. During training, the situations with high purity of values, i.e., high probability of winning or losing, are highlighted as the left nodes of the tree.

The training set is created based on the simulation results of 20 randomly created 3-year attacker-defender games. Only the results collected during the intelligent operation stage are used. The training set is initially separated into 4 subsets according to the attacker and defender roles and fleet types to highlight the differences due to modularity. The DTs for both fleets are shown in Fig. 2.

Based on these results, two important separation strategies are discovered that change the win-lose case: strategy 5 for the defender and strategy 2 for the attacker. Comparing Fig. 2.1 and Fig. 2.2, the modular fleet uses attack strategy 1 more frequent than the conventional fleet. The conventional fleet suffers a higher risk of losing when the modular fleet adopts attack strategy 1. Because the modular fleet makes more accurate prediction of the adversary, attack strategy 1 is more frequently used once a weak defense strategy of the conventional fleet is predicted.

Comparing Fig. 2.3 and Fig. 2.4, the modular fleet has a higher probability of winning when competing against a well-prepared adversarial defense convoy (46.9% vs. 33.9%). The modular fleet utilizes a strong attack strategy, i.e., $\geq$ attack strategy 4, to stay in the lead of the game. These results suggest that the higher win rate is due to the flexibility gained from fleet modularity. For the modular fleet, the idle transportation vehicles in the attack mission can be reconfigured to combat vehicles by swapping the capacity module with the weapon module to temporarily enhance the firepower of a convoy, which forms a stronger attack strategy.

## 4. Game Theoretical Model

Training the intelligent agent-based model is time-consuming. This increases the difficulty of proving the convergence and stability of the results. A multi-stage game model is created and fitted by using high-fidelity simulated data for performing a theoretical analysis. Each game is used to describe a single stage in the fleet competition. An example payoff matrix of a single-stage game is shown in Tab. 2.

**Table 2:** Payoff matrix for a single-stage game

|       | $d_1$   | $d_2$    | $d_3$    | $d_j$                         | $d_{10}$ |
|-------|---------|----------|----------|-------------------------------|----------|
| $a_1$ | (100,0) | (89,11)  | (47,53)  | ...                           | (0,100)  |
| $a_2$ | (100,0) | (98,2)   | (85,15)  | ...                           | (0,100)  |
| $a_3$ | (100,0) | (100,0)  | (90,10)  | ...                           | (0,100)  |
| $a_i$ | ...     | ...      | ...      | $(r_{m_i^1 c_j^1}, r_{c_j^1 m_i^1})$ | ...      |
| $a_{10}$ | (100,0) | (100,0) | (100,0)  | ...                           | (49,51)  |

Denote by $r_{m_i^t c_j^t}$ the payoff of the game that the modular fleet selects strategy $i$ and conventional fleet selects strategy $j$ in the $t^{th}$ stage. In this study, the payoffs are defined as the probabilities of wining, calculated by

$$r_{m_i^1 c_j^1} = \frac{n(\text{mod use } i \ \& \text{ conv use } i \ \& \ mod. win\,)}{n(\text{mod use } i \ \& \text{ conv use } j\,)}$$

where $n(X)$ is the number of samples in the training set that satisfy condition $X$.

The decision selected at the first stage may change the available decisions in games at the second stage due to the limitation in resources. For example, if the defender gives up the task at stage 1, i.e., if the defender choses strategy 1, then there is a high probability for the defender to perform a stronger strategy at stage 2, i.e., to choose one of the strategies 8, 9, or 10. However, if strategy 10 is used at stage 1, then strategy 1 may be the only available strategy at the stage 2. The payoffs for the unavailable strategies are zero, so availability of strategies changes the payoff matrix thus changing the game. Figure 3 shows an example of results obtained from the game model.

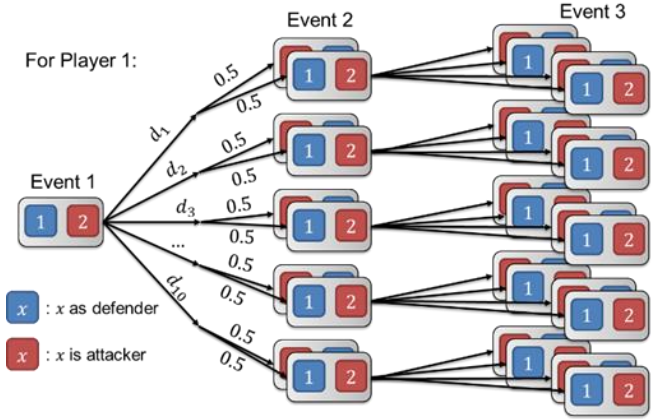

**Figure 3:** Three-stage game theoretical model with consideration the impact of previous decisions and assignment of attacker and defender.

Denote the payoff matrices of the possible game $k$ at stage 1 as $R_k^1$, $k \in K$, the payoff matrix of the modular fleet in the second game $R_{m_i^1}^2$ can be calculated as

$$R_{m_i^1}^2 = \sum_{k=1}^{k=K} \sum_{j=1}^{j=10} R_k^1 p(k|m_i^1, c_j^1)$$

where $p(k|m_i^1, c_j^1)$ is the probability of playing game $k$ at stage 2 given modular fleet selected strategy $i$ and given that the conventional fleet selected strategy $j$ at stage 1. By adding new conditions for the strategy selected at stage 2, we can also compute the payoff matrices for stage 3 as $R_{m_i^1, m_p^2}^3$.

Assume the impact of a decision can only last for three stages, a stochastic three-stage game can be formed to represent the attacker-defender game. Each payoff matrix is a 10 by 10 matrix. Vertex enumeration is applied to find the Nash equilibrium of this high-dimensional game [18]. To find the equilibrium strategy for a multi-stage game, we followed dynamic programming to solve the problem in a back-propagation manner. For this three-stage game, the procedures are:

1. Find the Nash equilibria and the corresponding payoffs of all possible games at stage 3.
2. Compute the payoff matrix that sums the payoffs at stage 2 and stage 3.
3. Find the strategy of stage 2 that leads to a Nash equilibrium for both stage 2 and stage 3.
4. Compute the payoff matrix that sums the payoffs at the first, second, and third stages.
5. Find the strategy at the first stage that will led to the Nash equilibrium for the first, second, and third stage.

A more detailed description of the algorithm is provided in Fig. 4. For a special case where the modular fleet plays as attacker-attacker-defender in three stages, the equilibrium strategies at different stages are shown in Tab. 3.

**Table 3:** Equilibrium strategy of game with different periods

|  | Fleet | Stage1 | Stage2 | Stage3 |
|---|---|---|---|---|
| Single Stage | Modular | $a10$ | N/A | N/A |
|  | Conv. | $d10$ | N/A | N/A |
| Two Stage | Modular | $a9$ | $a5$ | N/A |
|  | Conv. | $d10$ | $d5$ | N/A |
| Three Stage | Modular | $a8$ | $a2$ | $d10$ |
|  | Conv. | $d10$ | $d9$ | $a5$ |

The evolution of the equilibrium strategy is observed when changes in the number of stages of the game are considered. For a one-stage game, the equilibrium strategies are the strongest strategies for both the attacker and the defender. However, once the second stage is considered, the modular fleet saves part of the attack force at stage 1 for forming an attack convoy at stage 2. Once a third stage is considered, more attack force has been reserved for protecting the transportation convoy at stage 3 (last stage). With limited resources and consideration of risk of damage, each fleet needs to trade off the success rate in the current stage and the impact for operations at future stages.

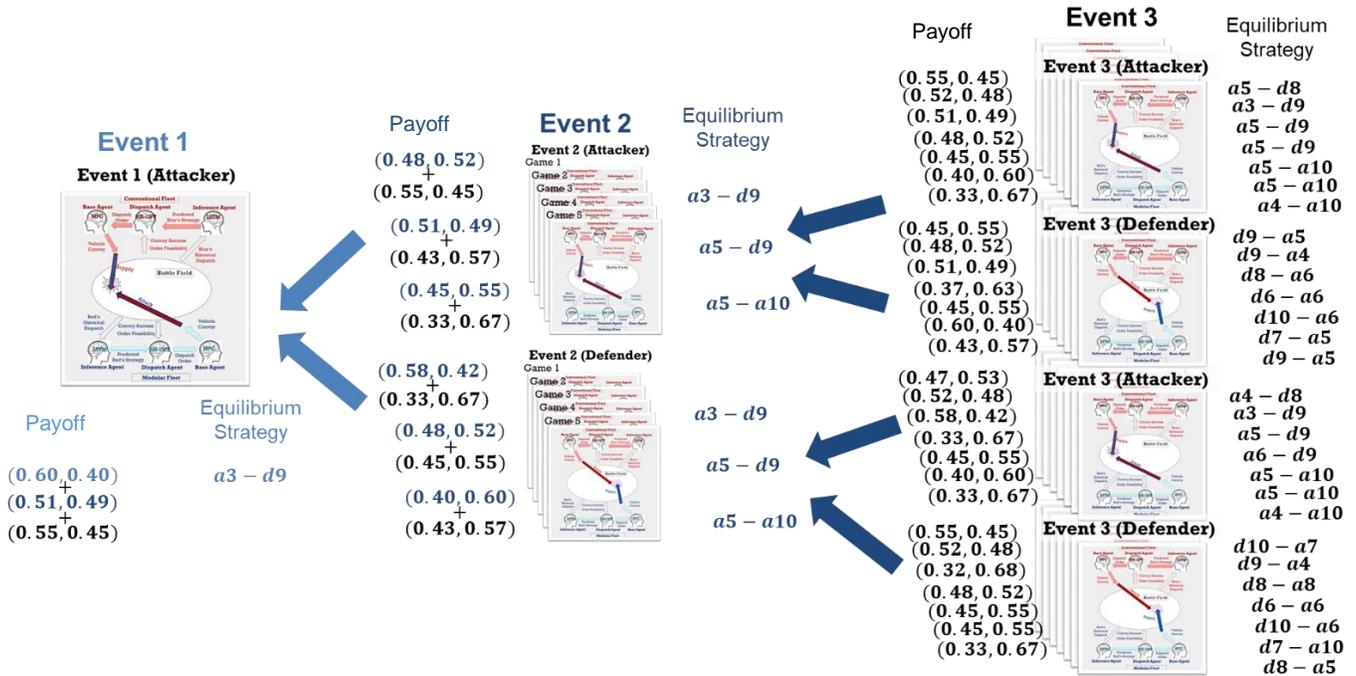

**Figure 4:** Flowchart to show the procedure of finding Nash equilibrium for a three-stage game.

## 5. Conclusions

In conclusion, this study provides a novel approach for evaluating the benefits and burdens of fleet modularity through the analysis of the competition between autonomous fleets in an attacker-defender game. An approach based on a decision tree was proposed for mining the operation heuristics and finding the main changes in fleet operation strategy once modularity is available. In addition, equilibria of operational strategy evolution over three stages were identified. The modular fleet outperforms the conventional fleet due to its additional flexibility in operation that makes it harder to predict by opponents. In the future, the stabilized win rate can be calculated by formulating an infinitely repeated game with proved convergence.

## ACKNOWLEDGMENTS

This research has been supported by the Automotive Research Center, a US Army Center of Excellence in Modeling and Simulation of Ground Vehicle Systems and headquartered at the University of Michigan. This support is gratefully acknowledged. The authors are solely responsible for the opinions contained herein.